\documentclass[runningheads]{llncs}
\usepackage{graphicx}
\usepackage{amsmath,amssymb} 
\usepackage{color}
\usepackage[width=122mm,left=12mm,paperwidth=146mm,height=193mm,top=12mm,paperheight=217mm]{geometry}

\usepackage[pagebackref=true,breaklinks=true,colorlinks,bookmarks=false]{hyperref}
\usepackage{times}
\usepackage{bm}
\usepackage{caption}
\usepackage{subfig}
\usepackage[nocompress]{cite}
\usepackage{widetable}
\usepackage{tabularx, booktabs}
\usepackage{cleveref}
\usepackage{sectsty}
\newcolumntype{Y}{>{\centering\arraybackslash}X}
\newcolumntype{C}{>{\centering\arraybackslash}p{9em}}

\begin{document}
\pagestyle{headings}
\mainmatter
\title{A 4D Light-Field Dataset and CNN Architectures for Material Recognition} 

\titlerunning{ }
\authorrunning{ }

\author{Ting-Chun Wang\inst{1} \and Jun-Yan Zhu\inst{1} \and Ebi Hiroaki\inst{2} \and \\ Manmohan Chandraker\inst{2} \and Alexei A. Efros\inst{1} \and Ravi Ramamoorthi\inst{2}}

\institute{University of California, Berkeley\\ \email{ \{tcwang0509,junyanz,efros\}@berkeley.edu} 
\and University of California, San Diego\\ \email { hebi@eng.ucsd.edu,\{mkchandraker,ravir\}@cs.ucsd.edu} }

\maketitle

\begin{abstract}
We introduce a new light-field dataset of materials, and take advantage of the recent success of deep learning to perform material recognition on the 4D light-field.
Our dataset contains 12 material categories, each with 100 images taken with a Lytro Illum, from which we extract about 30,000 patches in total.
To the best of our knowledge, this is the first mid-size dataset for light-field images.
Our main goal is to investigate whether the additional information in a light-field (such as multiple sub-aperture views and view-dependent reflectance effects) can aid material recognition. 
Since recognition networks have not been trained on 4D images before, we propose and compare several novel CNN architectures to train on light-field images.
In our experiments, the best performing CNN architecture achieves a 7\% boost compared with 2D image classification ($70\%\rightarrow 77\%$).
These results constitute important baselines that can spur further research in the use of CNNs for light-field applications. 
Upon publication, our dataset also enables other novel applications of light-fields, including object detection, image segmentation and view interpolation.

\end{abstract}

\section{Introduction}
Materials affect how we perceive objects in our daily life.
For example, we would not expect to feel the same when we sit on a wooden or leather chair.
However, differentiating materials in an image is difficult since their appearance depends on the confounding effects of object shape and lighting.
A more robust way to determine the material type is using the surface reflectance or the bidirectional reflectance distribution function (BRDF).
However, measuring the reflectance is hard.
Previous works use gonioreflectometers to recover the reflectance, which is cumbersome, and does not easily apply to spatially-varying BRDFs or Bidirectional Texture Functions (BTFs)~\cite{nicodemus1977geometrical, dana1999reflectance}.

An alternative to directly measuring the reflectance, is to consider multiple views of a point at once.
By doing so, material recognition can be improved as demonstrated by Zhang et al.~\cite{zhang2015reflectance}.
We exploit the multi-views in a light-field representation instead.
Light-field cameras have recently become available and are able to capture multiple viewpoints in a single shot.
We can therefore obtain the intensity variation under different viewing angles with minimal effort.
Therefore, one of the main goals of this paper is to investigate whether 4D light-field information improves the performance of material recognition over 2D images.
We adopt the popular convolutional neural network (CNN) framework to perform material classification in this work.
However, there are two key challenges:
First, all previous light-field datasets include only a few images, so they are not large enough to apply the data-hungry deep learning approaches.  
Second, CNN architectures have previously not been adapted to 4D light-fields;
Thus, novel architectures must be developed to perform deep learning with light-field inputs.  
Our contributions are shown in Fig.~\ref{fig:teaser} and summarized below: 

\begin{figure}[t]
\centering 
\includegraphics[width=.9\linewidth]{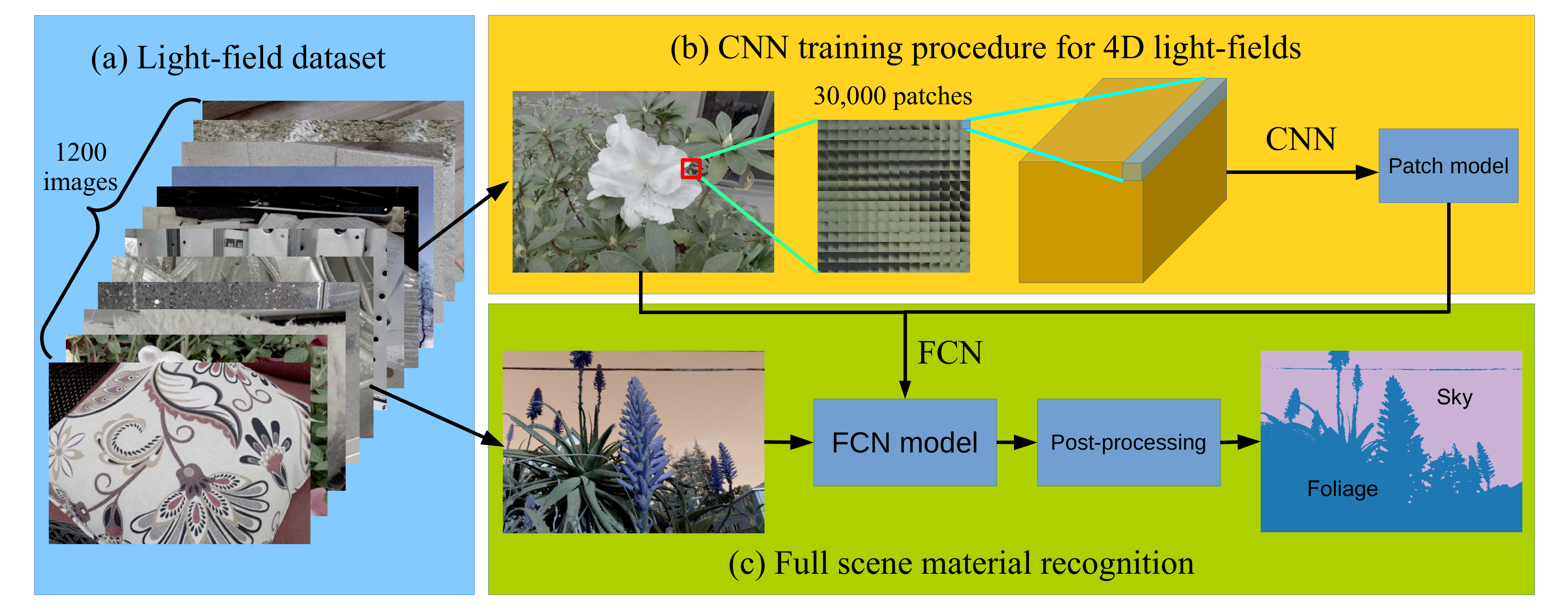}
\caption{\textit{Overview of our system and contributions. (a) We collect a new light-field dataset, which contains 1200 images labeled with 12 material classes. (b) Using (micro-lens) light-field patches extracted from this dataset, we train a CNN by modifying previous 2D models to take in 4D inputs. (c) Finally, we convert the patch model to an FCN model by fine-tuning on full images, and perform full scene material segmentation.}} 
\label{fig:teaser}
\end{figure}

\par \textbf{1)} We introduce the first mid-size light-field image dataset (Sec.~\ref{sec:dataset}).
Our dataset contains 12 classes, each with 100 images labeled with per pixel ground truth (Fig.~\ref{fig:dataset}).
We then extract 30,000 patches from these images.
Although we use this dataset for material recognition, it is not limited to this purpose and can be used for other light-field related applications.
Upon publication, the dataset will be released publicly.
\par \textbf{2)} We investigate several novel CNN architectures specifically designed for 4D light-field inputs (Sec.~\ref{sec:architecture}). 
Since no recognition CNN has been trained on light-fields before, we implement different architectures to work on 4D data (Figs.~\ref{fig:architecture} and~\ref{fig:filter}).
Instead of training a new network from scratch, we reuse the spatial filters from previous 2D models, while adding new angular filters into the network architecture.
We also find directly training a fully convolutional network (FCN) very unstable, and thus train on extracted patches first and fine-tune on full images afterwards.
The proposed architectures are not limited to material recognition, and may be used for other light-field based tasks as well.
\par \textbf{3)} Using our best-performing architecture, we achieve about 6-7\% boost compared with single 2D image material classification, increasing the accuracy from 70\% to 77\% on extracted patches and 74\% to 80\% on full images (Sec.~\ref{sec:exp}).
These act as important baselines for future work in light-field based material recognition.

\section{Related work}

\noindent{\bf Light-field datasets: }
The most popular dataset is the one introduced by Wanner et al.~\cite{wanner2013datasets}, which contains 7 synthetic scenes and 5 real images captured using a gantry.
Another well-known one is the Stanford light-field archive~\cite{stanford}, which provides around 20 light-fields sampled using a camera array, a gantry and a light-field microscope.
The synthetic light-field archive by Marwah et al.~\cite{marwah2013compressive} contains 5 camera light-fields and 13 display light-fields.
Other datasets contain fewer than ten images~\cite{tao2013depth, kim2013scene, jarabo2014people, wang2015occlusion} or are only suitable for particular purposes~\cite{li2014saliency,raghavendra2016exploring}.
Clearly, there is a lack of large light-field datasets in prior works.
In this work, we use the Lytro Illum camera to build a dataset with 1200 light-field images.

\noindent{\bf Material databases: }
The early work on material recognition was primarily on classifying instance-level textures, such as the CUReT database~\cite{dana1999reflectance} and the more diversified KTH-TIPS~\cite{hayman2004significance,caputo2005class} database.
Recently, the Describable Textures Dataset (DTD)~\cite{cimpoi2014describing} features real-world material images.
Some work on computer-generated synthetic datasets has also been introduced~\cite{li2012recognizing,weinmann2014material}.

For category-level material databases, the most well-known is the Flickr Material Database (FMD)~\cite{sharan2009material}, which contains ten categories with 100 images in each category. 
Subsequently, Bell et al.~\cite{bell2013opensurfaces} released OpenSurfaces which contains over 20,000 real-world scenes labeled with both materials and objects.  
More recently, the Materials in Context Database (MINC)~\cite{bell2015material} brought the data size to an even larger scale with 3 million patches classified into 23 materials.
However, these datasets are all limited to 2D, and thus unsuitable for investigating the advantages of using multiple views.
Although our dataset is not as large as the MINC dataset, it is the first mid-size 4D light-field dataset, and is an important step towards other learning based light-field research.
Zhang et al.~\cite{zhang2015reflectance} also propose a reflectance disk dataset which captures intensities of different viewing angles for 20 materials.
However, their dataset lacks the spatial information, and is much smaller compared to our dataset.

\noindent{\bf Material recognition: }
Material recognition methods can mainly be classified into two categories.
The first one recognizes materials based on the object reflectance ~\cite{cula20043d,lombardi2012single,liu2014discriminative,zhang2015reflectance}.
Most work of this type requires the scene geometry or illumination to be known, or requires special measurement of the BRDF beforehand.

The other body of work extracts features directly from the image appearance, and is thus more flexible and can work on real-world images.
Liu et al.~\cite{liu2010exploring} propose a model to combine low- and mid-level features using a Bayesian generative framework. 
Hu et al.~\cite{hu2011toward} extend the Kernel descriptors with variances of gradient orientations and magnitudes to handle materials.
Schwartz and Nishino~\cite{schwartz2013visual} introduce visual material traits and explicitly avoid object-specific information during classification.
Qi et al.~\cite{qi2014pairwise} introduce a pairwise transform invariant feature and apply it to perform material recognition.
Cimpoi et al.~\cite{cimpoi2014describing} propose a framework based on neural network descriptors and improved Fisher vectors (IFV).
Recently, Cimpoi et al.~\cite{cimpoi2015deep} combine object descriptors and texture descriptors to achieve state-of-the-art results on FMD.
However, none of these methods are applicable to the 4D case.
In this work, we implement different methods to deal with this dimensionality change from 2D to 4D.

\noindent{\bf Convolutional neural networks: }
Convolutional neural networks (CNNs) have proven to be successful in modern vision tasks such as detection and recognition, and are now the state-of-the art methods in most of these problems.
Since the work by Krizhevsky et al.~\cite{krizhevsky2012imagenet} (a.k.a. AlexNet), in recent years many advanced architectures have been introduced, including GoogLeNet~\cite{szegedy2015going} and VGG~\cite{simonyan2014very}.
For per-pixel segmentation, Farabet et al.~\cite{farabet2013learning} employ a multi-scale CNN to make class predictions at every pixel in a segmentation. 
A sliding window approach is adopted by Oquab et al.~\cite{oquab2014learning} to localize patch classification of objects. 
Recently, a fully convolutional framework~\cite{long2015fully} has been proposed to generate dense predictions from an image directly.

\noindent{\bf Multi-image CNNs: }
For CNNs trained on multiple image inputs, Yoon et al.~\cite{yoon2015learning} train a super-resolution network on light-field images; however, their goal is different from a high-level recognition task.
Besides, only a couple of images instead of the full light-fields are sent into the network at a time, so the entire potential of the data is not exploited.
Su et al.~\cite{su2015multi} propose a ``viewpooling'' framework to combine multiple views of an object to perform object recognition.
In their architecture, convolutional maps independently extracted from each view are maxpooled across all views. 
However, we find this does not work well in the light-field case. 
Rather, we demonstrate that it is advantageous to exploit the structure of light-fields in combining views much earlier in the network.
This also has the advantage that memory usage is reduced.
In this work, to ease the training of 4D light-fields, we initialize the weights with pre-trained 2D image models.
We investigate different ways to map the 4D light-field onto the 2D CNN architecture, which has not been explored in previous work, and may be beneficial to learning-based methods for other light-field tasks in the future.

\section{The light-field material dataset} \label{sec:dataset}

While the Internet is abundant with 2D data, light-field images are rarely available online.
Therefore, we capture the images ourselves using the Lytro Illum camera.
There are 12 classes in our dataset: fabric, foliage, fur, glass, leather, metal, plastic, paper, sky, stone, water, and wood.
Each class has 100 images labeled with material types.
Compared with FMD~\cite{sharan2009material}, we add two more classes, fur and sky.
We believe these two classes are very common in natural scenes, and cannot be easily classified into any of the ten categories in FMD.

The images in our dataset are acquired by different authors, in different locations (e.g.\ shops, campus, national parks), under different viewpoints and lighting conditions, and using different camera parameters (exposure, ISO, etc).
The spatial resolution of the images is $376\times 541$, and the angular resolution is $14\times 14$.
Since the pixel size of the Lytro camera is small ($1.4 \mu m$), one problem we encountered is that the images are often too dark to be usable.
Water is also a particularly challenging class to capture, since the corresponding scenes usually entail large motions.
Overall, of the 1448 acquired images, we retain 1200 not deemed too repetitive, dim or blurred.
We then manually classified and labeled the images with per pixel material category using the \emph{Quick Selection Tool} of Photoshop. 
For each material region, we manually draw the boundary along the region. 
We check the segmentation results, and further refine the boundaries until we obtain final accurate annotation.

\begin{figure}[t]
\centering 
\includegraphics[width=.9\linewidth]{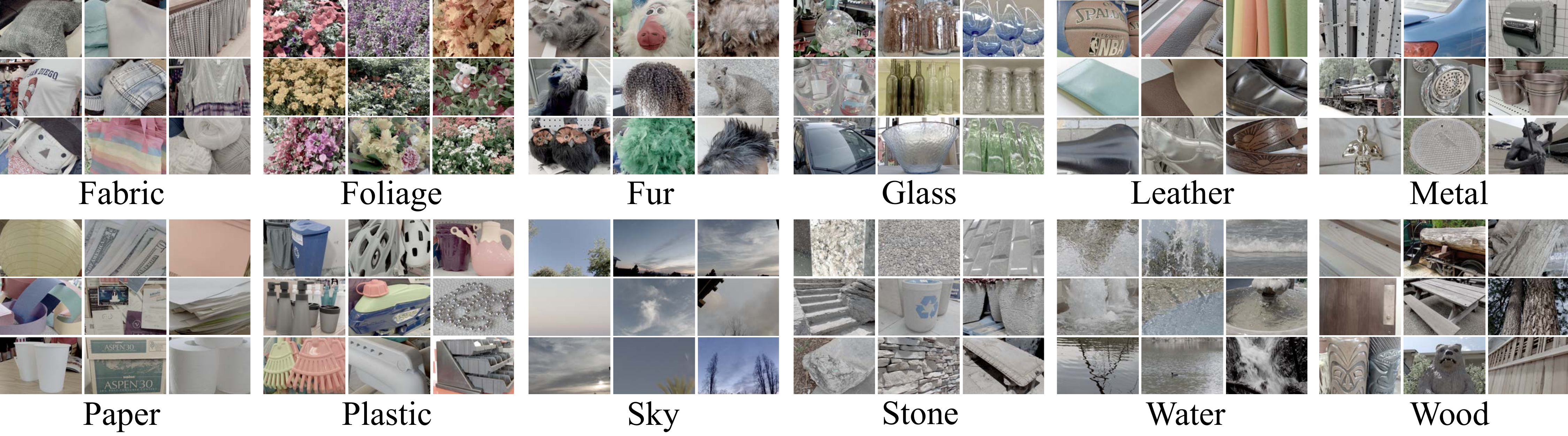}
\caption{\textit{Example images in our dataset. Each class contains 100 images.}}
\label{fig:dataset}
\end{figure}

\begin{figure}[t]
\centering 
\null\hfill
\subfloat[2D vs. LF images]{\includegraphics[width=.48\linewidth]{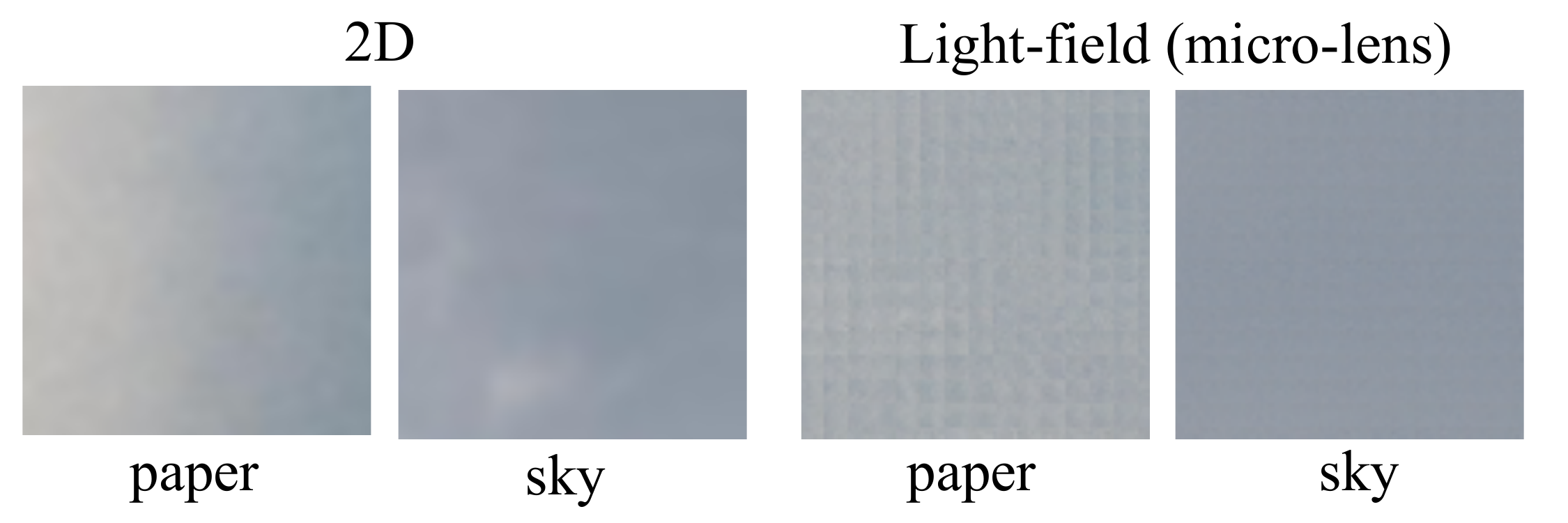} \label{fig:benefit:a}} \hfill
\subfloat[2D vs. LF predictions]{\includegraphics[width=.46\linewidth]{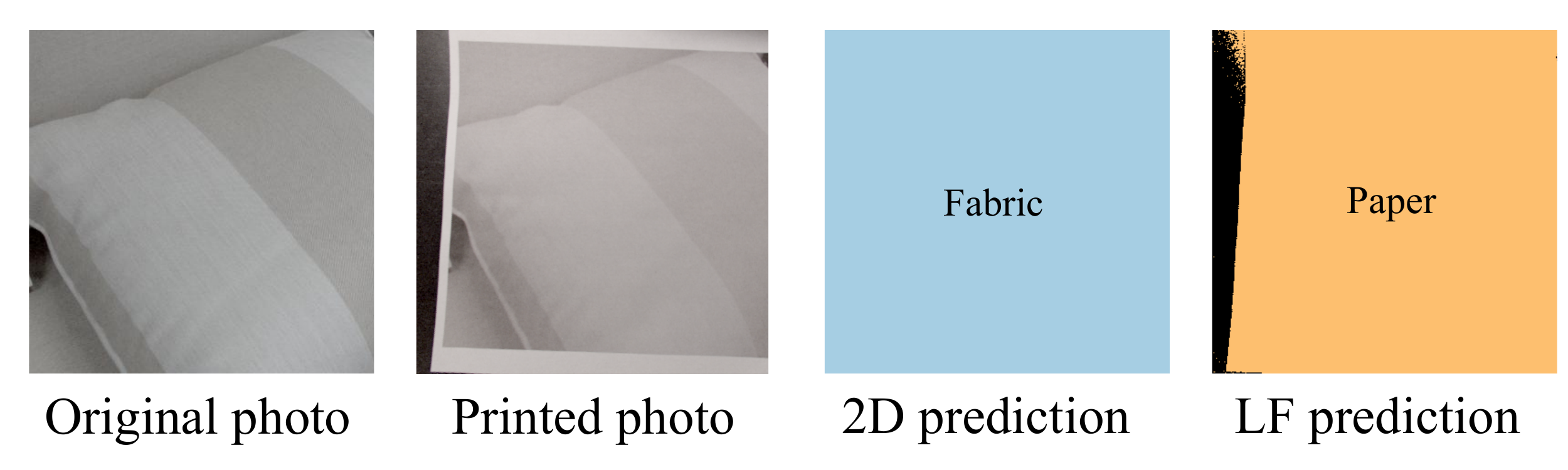} \label{fig:benefit:b}} \hfill\null
\caption{\textit{Example benefits of using light-field images. (a) From the 2D images, it is difficult to distinguish between paper and sky. However, with the light-field images it becomes much easier.
(b) We print out a picture of a pillow, and test both 2D and light-field models on the picture. The 2D model, without any reflectance information, predicts the material as fabric, while the light-field model correctly identifies the material as paper.}} 
\label{fig:benefit}
\end{figure}

In Fig.~\ref{fig:dataset} we show some example images for each category of the dataset.
Then, in Fig.~\ref{fig:benefit:a} we show example light-field images, where each block of pixels shows different viewpoints of a 3D point.
We then demonstrate the benefits of using light-fields: 
from the 2D images alone, it is difficult to separate sky from blue paper due to their similar appearances;
However, with the aid from light-field images, it becomes much easier since paper has different reflectance from different viewpoints while sky does not.
Next, in Fig.~\ref{fig:benefit:b} we print out a photo of a pillow, and take a picture of the printed photo.
We then test both 2D and light-field models on the picture.
It is observed that the 2D model predicts the material as fabric since it assumes it sees a pillow, while the light-field model correctly identifies the material as paper.

Finally, to classify a point in an image, we must decide the amount of surrounding context to include, that is, determine the patch size.
Intuitively, using small patches will lead to better spatial resolution for full scene material segmentation, but large patches contain more context, often resulting in better performance.
Bell et al.~\cite{bell2015material} choose the patch scale as $23.3\%$ of the smaller image length, although they find that scale 32\% has the best performance.
Since our images are usually taken closer to the objects, we use $34\%$ of the smaller image length as the patch size, which generates about 30,000 patches of size $128\times 128$.
This is roughly 2500 patches in each class.
The patch centers are separated by at least half the patch size; also, the target material type occupies at least half the patch.
Throughout our experiments, we use an angular resolution of $7\times 7$.
We randomly select 70\% of the dataset as training set and the rest as test set.
Patches from the same image are either all in training or in test set, to ensure that no similar patches appear in both training and test sets.

\section{CNN architectures for 4D light-fields} \label{sec:architecture}
\begin{figure}[t!]
\centering 
\null\hfill
\subfloat[viewpool]{\includegraphics[width=.4\linewidth]{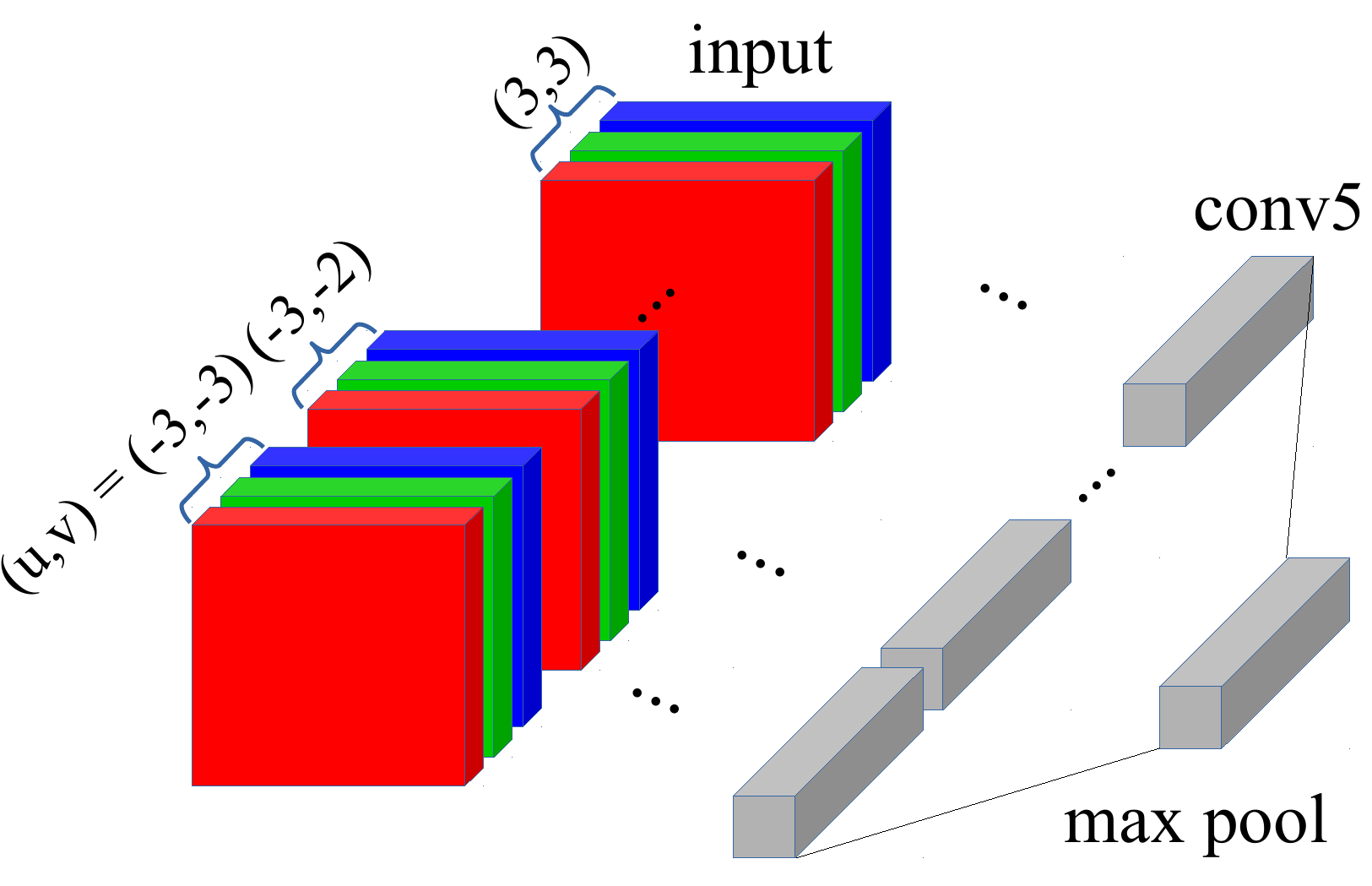} \label{fig:architecture:a}} \hfill
\subfloat[stack]{\includegraphics[width=.4\linewidth]{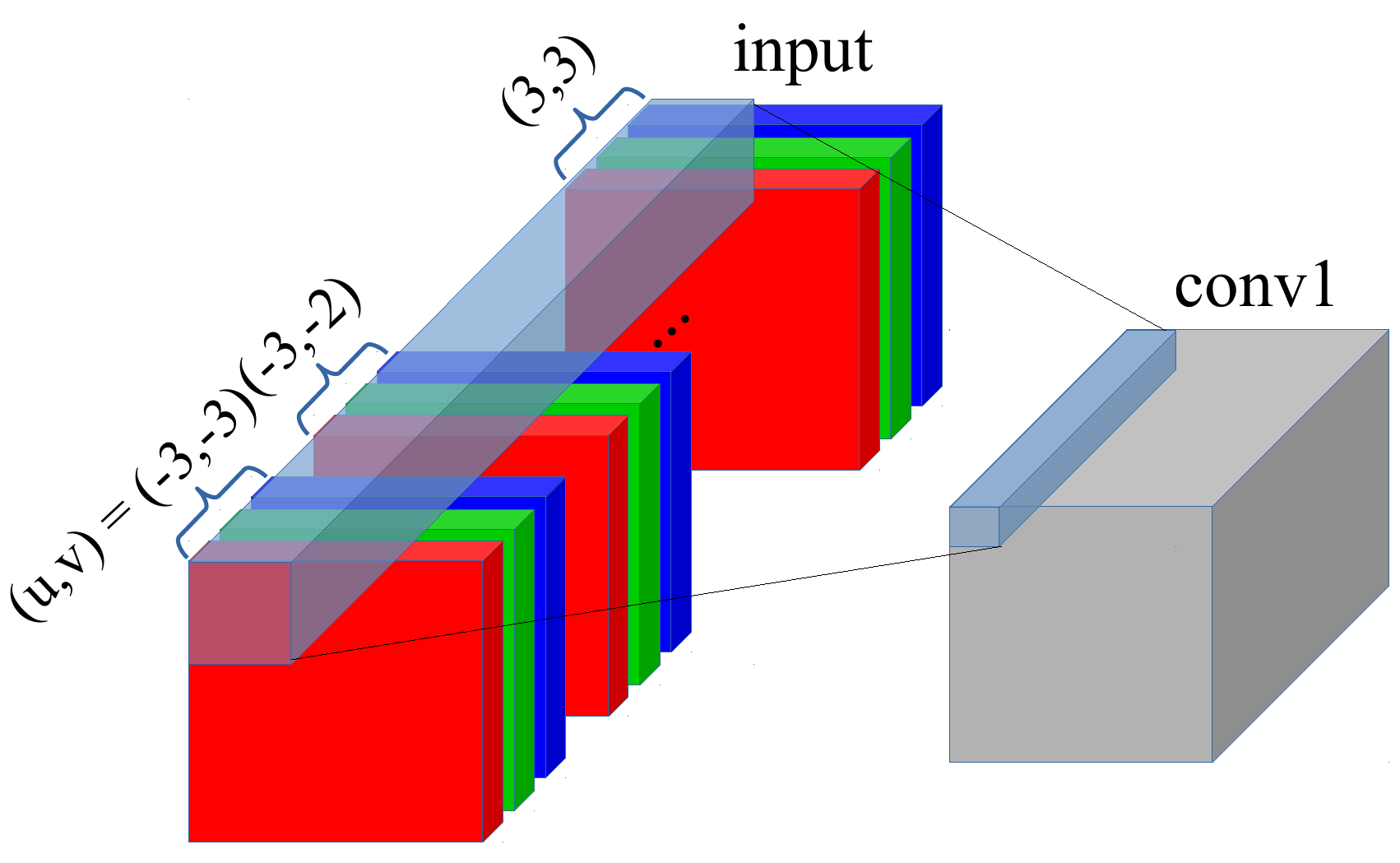} \label{fig:architecture:b}} \hfill\null \\
\null\hfill
\subfloat[EPI]{\includegraphics[width=.4\linewidth]{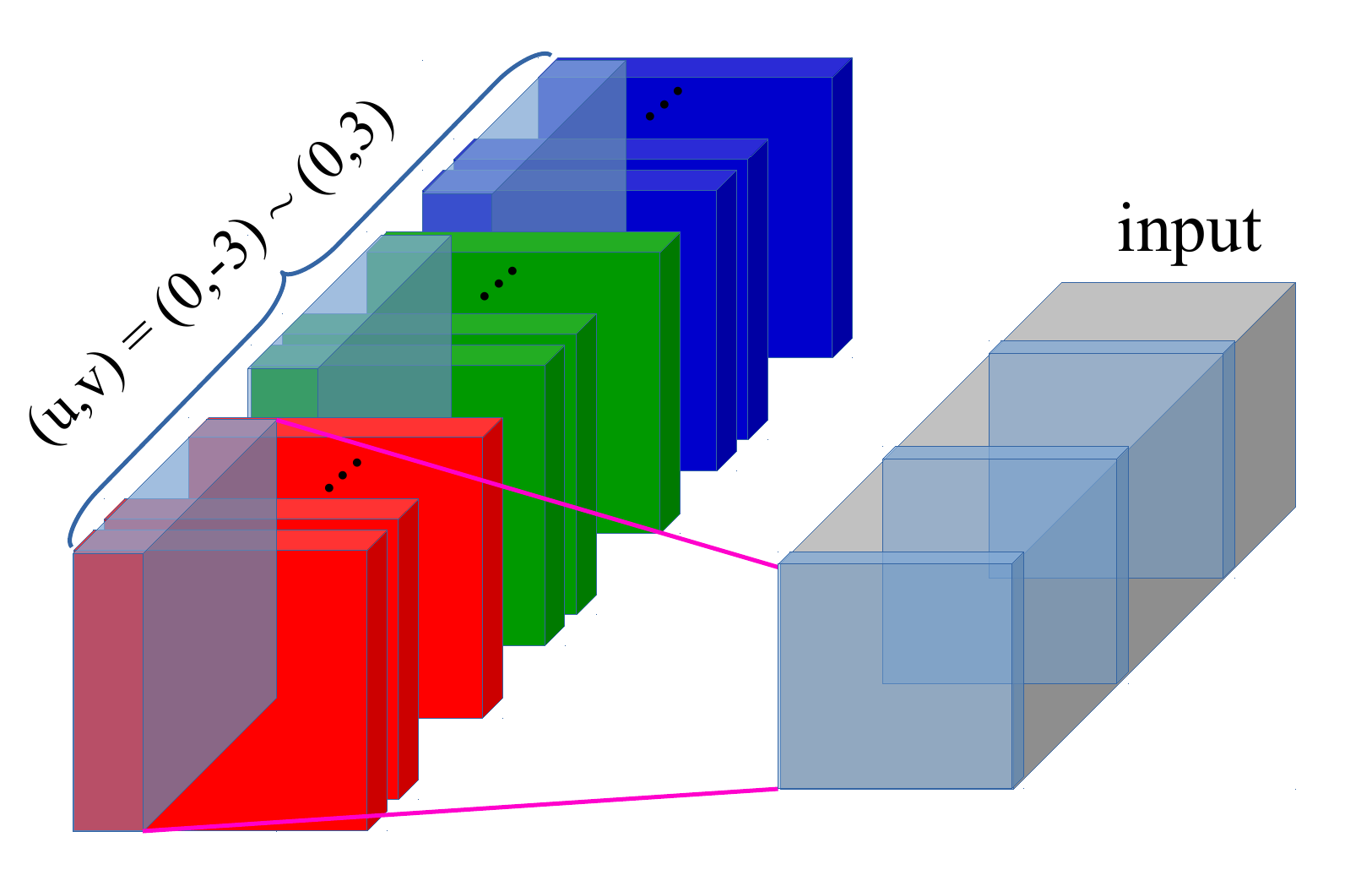}\label{fig:architecture:c}} \hfill
\subfloat[angular filter]{\includegraphics[width=.4\linewidth]{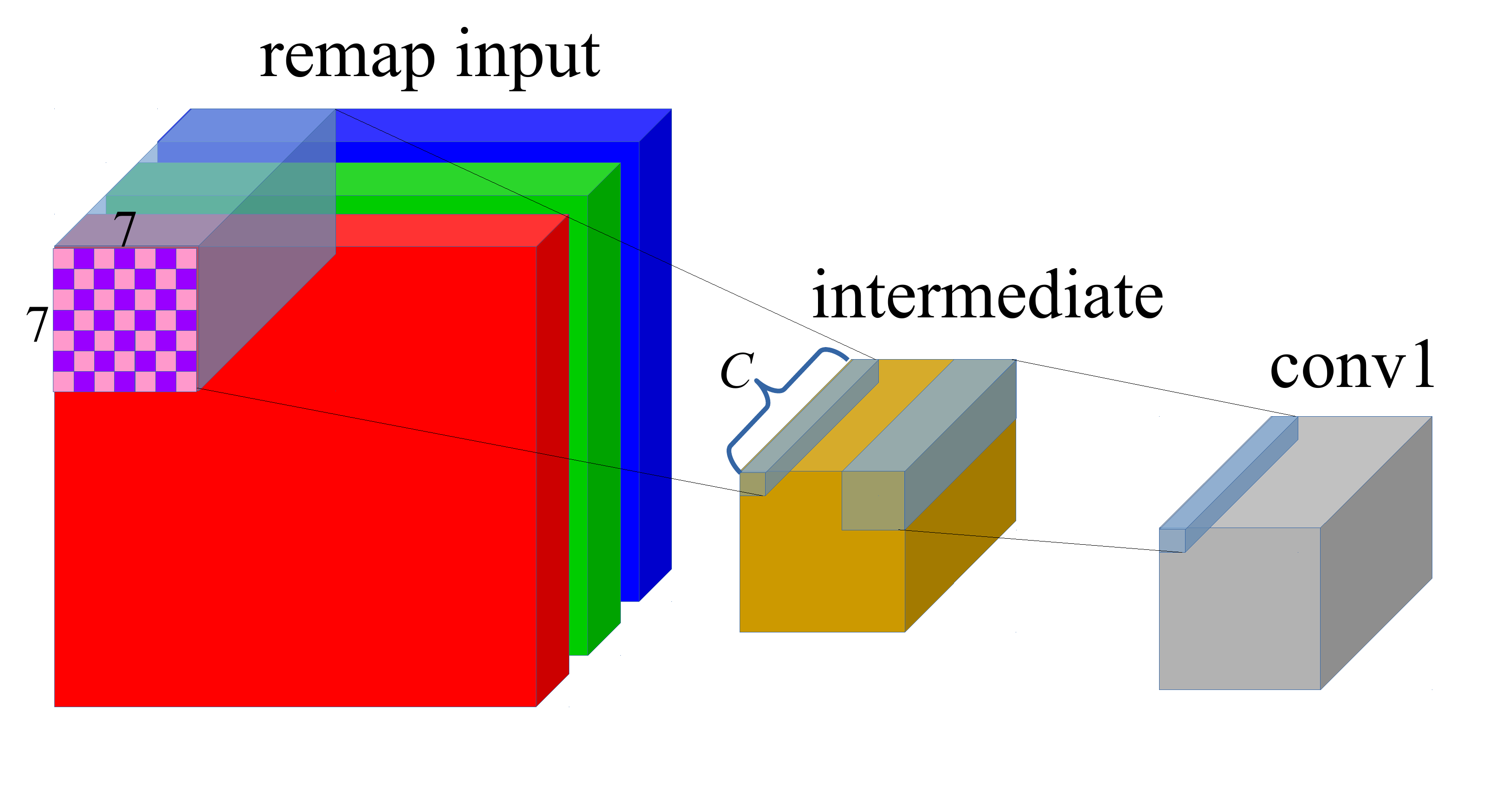}\label{fig:architecture:d}} \hfill\null
\caption{\textit{Different CNN architectures for 4D light-field inputs. The RGB colors represent the RGB channels, while $(u,v)$ denotes different angular coordinates (from $(-3,-3)$ to $(3,3)$ in our experiments). 
(a) After each view is passed through the convolutional part, they are max pooled and combined into one view, and then sent to the fully connected part. 
(b) All views are stacked across the RGB channels to form the input.
(c) The inputs are the horizontal and vertical EPIs concatenated together (only vertical ones are shown in the figure). 
(d) A $7\times 7$ angular filter is first applied on the remap image. The intermediate output is then passed to the rest of the network. 
}}
\label{fig:architecture}
\end{figure}

We now consider the problem of material recognition on our 4D light-field dataset and draw contrasts with recognition using 2D images.
We train a Convolutional Neural Network for this patch classification problem.
Formally, our CNN is a function $f$ that takes a light-field image $R$ as input and outputs a confidence score $p_k$ for each material class $k$.
The actual output of $f$ depends on the parameters $\theta$ of the network that are tuned during training, i.e., $p_k = f(R;\theta)$.
We adopt the softmax loss, which means the final loss for a training instance is $-\log (e^{p_t}/(\sum_{i=1}^k e^{p_i}))$, where $t$ is the true label.
At test time, we apply the softmax function on the output $p_k$, where the results can be seen as the predicted probability per class.

We use the network architecture of the recent VGG-16 model~\cite{simonyan2014very}, a 16-layer model, as it performs the best on our dataset when using 2D images.
We initialize the weights using the MINC VGG model~\cite{bell2015material}, the state-of-the-art 2D material recognition model, and then fine-tune it on our dataset. 

The biggest challenge, however, is we have 4D data instead of 2D.
In other words, we need to find good representations for 4D light-field images that are compatible with 2D CNN models.
We thus implement a number of different architectures and report their performance.
The results may be used as baselines, and might be useful for designing other learning-based methods for light-fields in the future.
In our implementation, the best performing methods (angular filter and 4D filter) achieve 77\% classification accuracy on the extracted patches, which is 7\% higher than using 2D images only.
Details of each architecture are described below.

\noindent{\bf 2D average } First and the simplest, we input each image independently and average the results across different views.
This, however, definitely does not exploit the implicit information inside light-field images.
It is also time consuming, where the time complexity grows linearly with angular size.

\noindent{\bf Viewpool } Second, we leverage the recent ``viewpool'' method proposed in~\cite{su2015multi} (Fig.~\ref{fig:architecture:a}).
First, each view is passed through the convolutional part of the network separately. 
Next, they are aggregated at a max view-pooling layer, and then sent through the remaining (fully-connected) part of the network.
This method combines information from different views at a higher level; however, max pooling only selects one input, so still only one view is chosen at each pixel.
Also, since all views need to be passed through the first part of the network, the memory consumption becomes extremely large.

\noindent{\bf Stack } Here, we stack all different views across their RGB channels before feeding them into the network, and change only the input channel of the first layer while leaving the rest of the architecture unchanged (Fig.~\ref{fig:architecture:b}).
This has the advantage that all views are combined earlier and thus takes far less memory.

\noindent{\bf EPI } For this method, we first extract the horizontal and vertical epipolar images (EPIs) for each row or column of the input light-field image. 
In other words, suppose the original 4D light-field is $L(x,y,u,v)$, where $(x,y)$ are the spatial coordinates and $(u,v)$ are the angular coordinates. We then extract 2D images from $L$ by
\begin{equation}
\begin{split}
L(x,y=y_i,u,v=0) \quad \forall i=1,...,h_s \\
L(x=x_j,y,u=0,v) \quad \forall j=1,...,w_s
\end{split}
\end{equation} 
where $(u,v)=(0,0)$ is the central view and $(h_s,w_s)$ are the spatial size.
These EPIs are then concatenated into a long cube and passed into the network (Fig.~\ref{fig:architecture:c}).
Again only the first layer of the pre-trained model is modified.

\noindent{\bf Angular filter on remap image} 
The idea of applying filters on angular images was first proposed by Zhang et al.~\cite{zhang2015reflectance}.
However, they only considered 2D angular images, while in our case we have 4D light-field images.
Also, by incorporating the filters into the neural network, we can let the network learn the filters instead of manually designing them, which should achieve better performance.

For this method, we use the remap image instead of the standard image as input. 
A remap image replaces each pixel in a traditional 2D image with a block of angular pixels $h_a \times w_a$ from different views, where $(h_a,w_a)$ are the angular size. 
The remap image is thus of size $(h_a\times h_s)\times(w_a\times w_s)$. 
It is also similar to the raw micro-lens image the Lytro camera captures; the only difference is that we eliminate the boundary viewpoints where the viewing angles are very oblique.

Before sending the remap image into the pre-trained network, we apply on it an angular filter of size $h_a\times w_a$ with stride $h_a,w_a$ and output channel number $C$ (Fig.~\ref{fig:architecture:d}).
After passing this layer, the image reduces to the same spatial size as the original 2D input.
Specifically, let this layer be termed $I$ (intermediate), then the output of this layer for each spatial coordinate $(x,y)$ and channel $j$ is
\begin{equation}
\ell^{j}(x,y) = g\Big(\sum_{i=r,g,b}\sum_{u,v} w_{i}^{j}(u,v) L^{i}(x,y,u,v)\Big) \quad \forall j = 1,...,C
\end{equation}
where $L$ is the input (RGB) light-field, $i,j$ are the channels for input light-field and layer $I$, $w_{i}^{j}(u,v)$ are the weights of the angular filter, and $g$ is the rectified linear unit (ReLU).
Afterwards, $\ell^{j}$ is passed into the pre-trained network.

\begin{figure}[t!]
\centering 
\null\hfill
\subfloat[spatial filter]{\includegraphics[width=.23\linewidth]{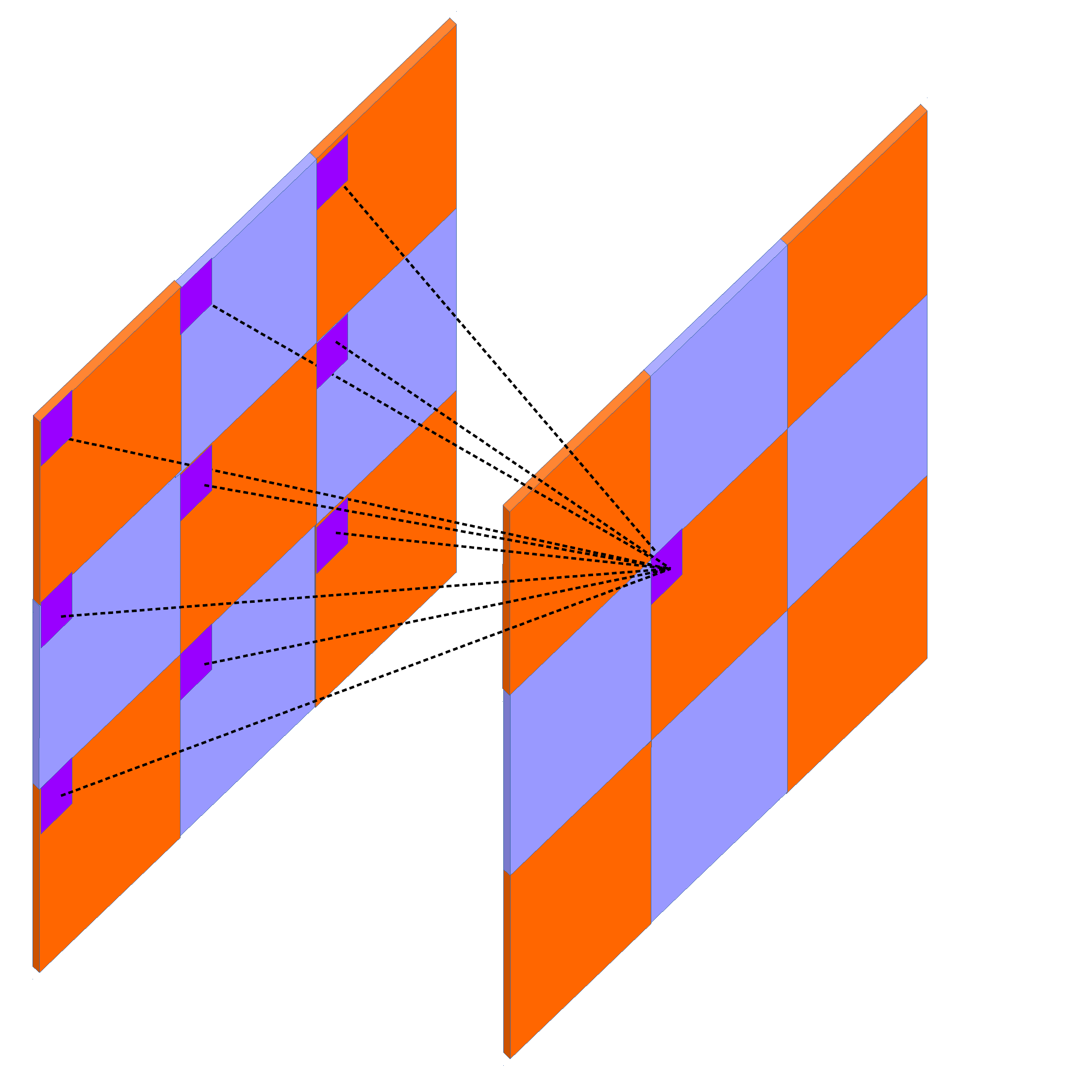} \label{fig:filter:a}} \hfill
\subfloat[angular filter]{\includegraphics[width=.23\linewidth]{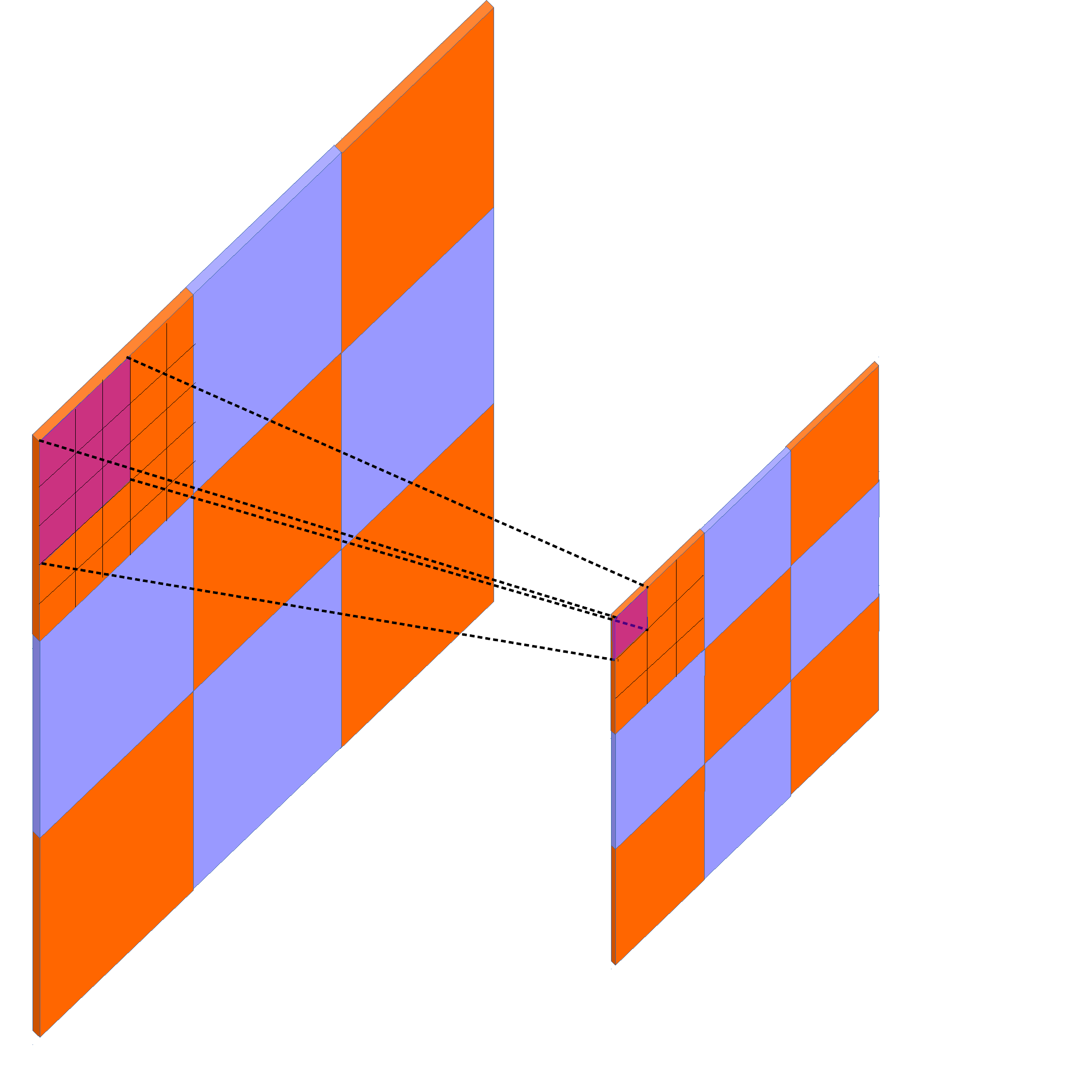} \label{fig:filter:b}} \hfill
\subfloat[interleaved filter]{\includegraphics[width=.25\linewidth]{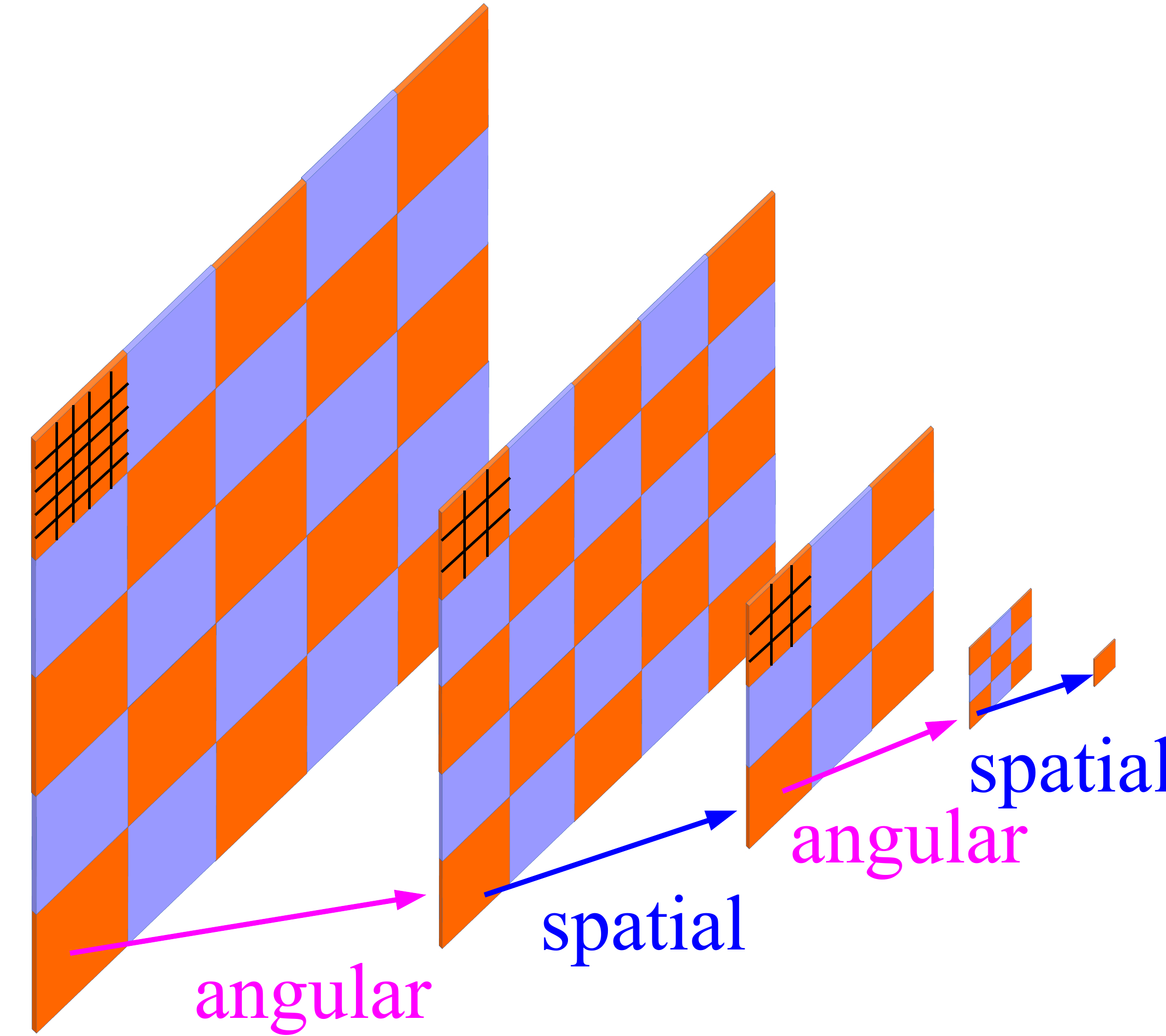} \label{fig:filter:c}} \hfill\null
\caption{\textit{(a)(b) New spatial and angular filters on a remap light-field image. The pooling feature is also implemented in a similar way. (c) By interleaving the angular and spatial filters (or vice versa), we mimic the structure of a 4D filter. }}
\label{fig:filter}
\end{figure}

\noindent{\bf 4D filter } Finally, since the light-field has a 4D structure, it becomes intuitive to apply a 4D filter to train the network.
However, directly applying a 4D filter is problematic due to several reasons.
First, a 4D filter contains far more parameters than a 2D filter. 
Even the smallest 4D filter ($3\times 3\times 3\times 3$) contains the same number of parameters as a $9\times 9$ 2D filter.
This is expensive in terms of both computation and memory.
Second, a 4D filter is not present in any pre-trained network, so we need to train it from scratch, which means we cannot take advantage of the pre-trained model.

Our solution is to decompose a 4D filter into two consecutive 2D filters, a spatial filter and an angular filter, implemented on a remap image as shown in Fig.~\ref{fig:filter}.
The new spatial filter is similar to a traditional 2D filter, except that since we now work on ``blocks'' of pixels, it takes only one pixel from each block as input (Fig.~\ref{fig:filter:a}).
This can be considered as a kind of ``stride'', but instead of stride in the output domain (where the filter itself moves), we have stride in the input domain (where the input moves while the filter stays).
The angular filter, on the other hand, convolves an internal block normally just as a traditional 2D filter, but does not work across the block boundaries (Fig.~\ref{fig:filter:b}).
By interleaving these two types of filters, we can approximate the effect of a 4D filter while not sacrificing the advantages stated above (Fig.~\ref{fig:filter:c}).
The corresponding pooling structures are also implemented in the same way.
To use the pre-trained models, the parameters from the original spatial filters are copied to the new spatial filters, while the angular filters are inserted between them and trained from scratch.

\section{Experimental results} \label{sec:exp}

The various architectures in Sec.~\ref{sec:architecture} are trained end-to-end using back-propagation.
To ease the training of 4D light-fields, we initialize the weights with pre-trained 2D image models.
The optimization is done with Stochastic Gradient Descent (SGD) using the Caffe toolbox~\cite{jia2014caffe}.
The inputs are patches of spatial resolution $128 \times 128$ and angular resolution $7 \times 7$.
To bring the spatial resolution to the normal size of $256 \times 256$ for VGG, we add a deconvolution layer at the beginning.
We use a basic learning rate of $10^{-4}$, while the layers that are modified or newly added use 10 times the basic learning rate.
Below, we present a detailed performance comparison between different scenarios.

\subsection{Comparison of different CNN architectures}
We first compare the prediction accuracies for different architectures introduced in the previous section.
Each method is tested 5 times on different randomly divided training and test sets to compute the performance average and variance.
In Table~\ref{tab:arch_acc}, the first column (2D) is the result of the MINC VGG-16 model fine-tuned on our dataset, using only a single (central view) image.
The remaining columns summarize the results of the other 4D architectures.
Note that these 4D methods use more data as input than a 2D image; we will make a comparison where the methods take in an equal number of pixels in Sec.~\ref{sec:exp:spa}, and the results are still similar.

As predicted, averaging results from each view (2D avg) is only slightly better than using a 2D image alone.
Next, the viewpool method actually performs slightly worse than using a 2D input;
this indicates that the method is not suitable for light-fields, where the viewpoint changes are usually very small.
The stack method and the EPI method achieve somewhat better performance, improving upon 2D inputs by 2-3\%.
The angular filter method achieves significant improvement over other methods; compared to using 2D input, it obtains about 7\% gain.
This shows the advantages of using light-fields rather than 2D images, as well as the importance of choosing the appropriate representation.
The 4D filter method achieves approximately the same performance as the angular filter method.
However, the angular filter method consumes much less memory, so it will be used as the primary comparison method in the following.
The performance of each material class for the angular filter method is detailed in Table~\ref{tab: lf_gain}.

\begin{table}[t]
\begin{center}
\begin{tabularx}{\textwidth}{|c *{8}{|Y}|} \hline
Architecture & 2D & 2D avg & viewpool & stack & EPI & angular & 4D \\    \hline
Accuracy (\%) &  $70.2_{\pm1.0}$  &  $70.5_{\pm0.9}$  &  $70.0_{\pm1.0}$  &  $72.8_{\pm1.1}$  &  $72.3_{\pm1.0}$  &  $\mathbf{77.0_{\pm1.1}}$ &  $\mathbf{77.0_{\pm1.1}}$ \\ \hline
\end{tabularx}
\end{center}
\caption{\textit{Classification accuracy (average and variance) for different architectures. The 2D average method is only slightly better than using a single 2D image; the viewpool method actually performs slightly worse. The stack method and the EPI method both achieve better results. Finally, the angular filter method and the 4D filter method obtain the highest accuracy.}}
\label{tab:arch_acc}
\end{table}

\begin{table}[t]
\begin{center}
\begin{tabular}{|*{4}{C|}} \hline
Fabric: 65.5\%  & Foliage: 92.5\% & Fur: 77.9\% & Glass: 65.2\% \\ \hline 
Leather: 91.1\% & Metal: 73.5\%& Paper: 60.4\% & Plastic: 50.0\% \\ \hline 
Sky: 98.2\% & Stone: 87.1\% & Water: 92.0\% & Wood: 72.6\% \\ \hline 
\end{tabular}
\end{center}
\caption{\textit{Patch accuracy by category for the angular filter method. }}
\label{tab: lf_gain}
\end{table}

To further test the angular filter method, we compare performances by varying three parameters: the filter location, the filter size, and the number of output channels of the angular filter.
First, we apply the angular filter at different layers of the VGG-16 network, and compare their performance.
The classification accuracies when the filter is applied on layer 1 and layer 2 are $76.6\%$ and $73.7\%$, respectively.
Compared with applying it on the input directly ($77.8\%$), we can see that the performance is better when we combine different views earlier.
This also agrees with our findings on the viewpool method and 4D method.
Next, we decompose the $7\times 7$ angular filter into smaller filters.
The accuracies for three consecutive $3\times 3$ filters and a $5\times 5$ filter followed by a $3\times 3$ filter are $74.8\%$ and $73.6\%$, respectively. 
It can be seen that making the filters smaller does not help improve the performance.
One reason might be that in contrast to the spatial domain, where the object location is not important, in the angular domain the location actually matters (e.g. the upper-left pixel has a different meaning from the lower-right pixel), so a larger filter can better capture this information.
Finally, we vary the number of output channels of the angular filter.
Since the filter is directly applied on the light-field input, this can be considered as a ``compression'' of the input light-field.
The fewer channels we output, the more compression we achieve using these filters.
We test the number from 3 (all views compressed into one view) to 147 (no compression is made), and show the results in Table~\ref{tab: remap_channel}.
It can be seen that the performance has a peak at 64 channels. 
We hypothesize that with fewer channels, the output might not be descriptive enough to capture variations in our data, but a much larger number of channels leads to overfitting due to the resulting increase in number of parameters.

\begin{table}[t]
\begin{center}
\begin{tabularx}{0.85\textwidth}{|c *{7}{|Y}|} \hline
Number of channels & 3 & 16 & 32 & 64 & 128 & 147 \\    \hline
Accuracy  &  71.6\%  &  74.8\%  &  76.7\%  &  \textbf{\emph{77.8\%}}  &  73.6\%  &  72.8\% \\ \hline
\end{tabularx}
\end{center}
\caption{\textit{Number of output channels of the angular filter architecture. As we can see, using more channels increases the performance up to some point (64 channels), then the performance begins to drop, probably due to overfitting.
This may also be related to light-field compression, where we do not need the entire 49$\times$3 input channels and can represent them in fewer channels for certain purposes.}}
\label{tab: remap_channel}
\end{table}

\subsection{Comparison between 2D and light-field results}
\begin{figure}[t]
\centering 
\null\hfill
\subfloat[2D]{\includegraphics[width=.4\linewidth]{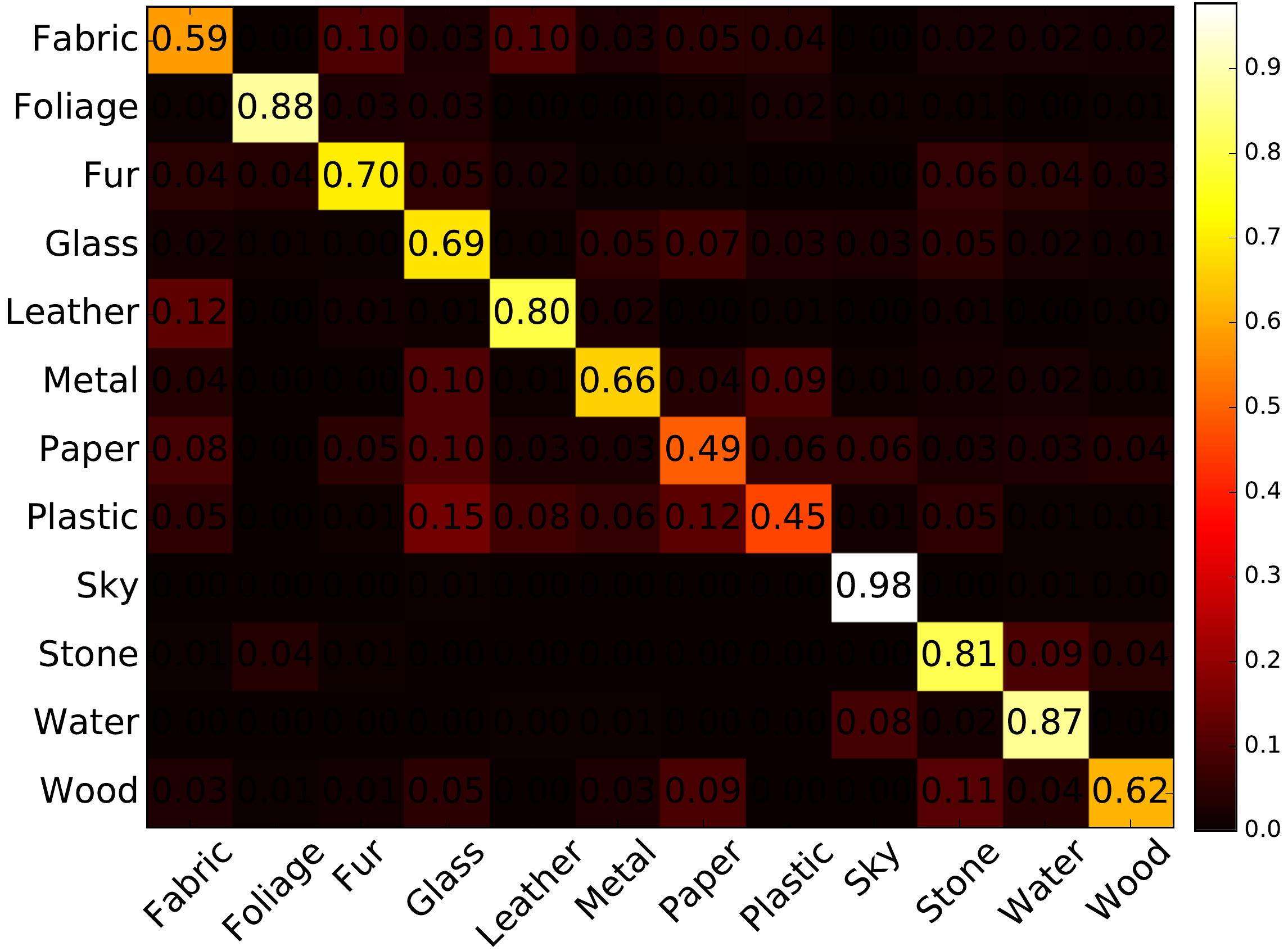} }\hfill
\subfloat[light-field]{\includegraphics[width=.4\linewidth]{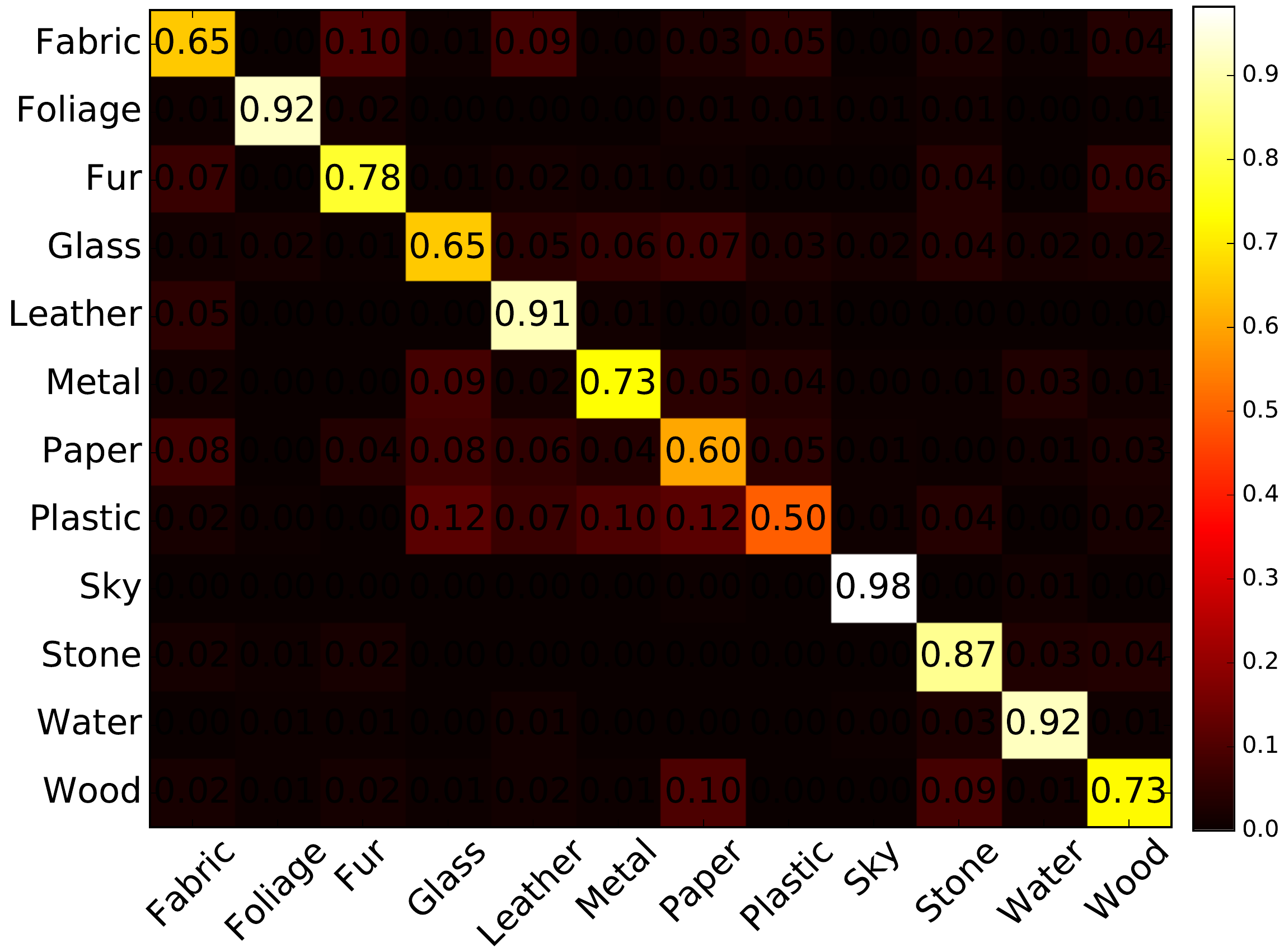} } \hfill\null
\caption{\textit{Confusion matrix comparison between 2D and light-field results.}}
\label{fig:confusion}
\end{figure}

\begin{figure}[t!]
\centering 
\null\hfill
\subfloat[accuracy by category]{\includegraphics[width=.42\linewidth,height=0.143\paperheight]{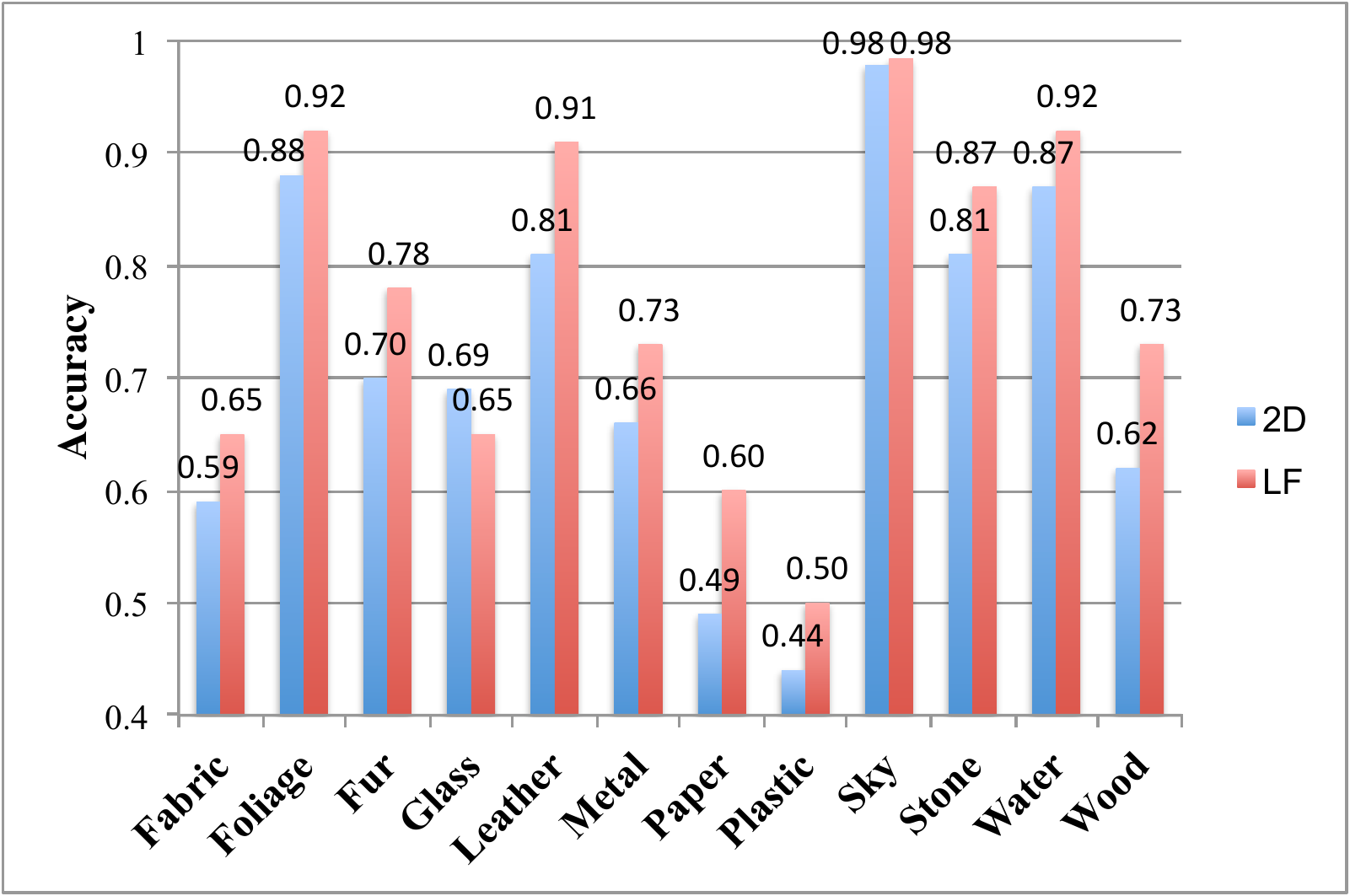}\label{fig: lf_gain:a}} \hfill
\subfloat[accuracy vs. patch sizes]{\includegraphics[width=.42\linewidth,height=0.143\paperheight]{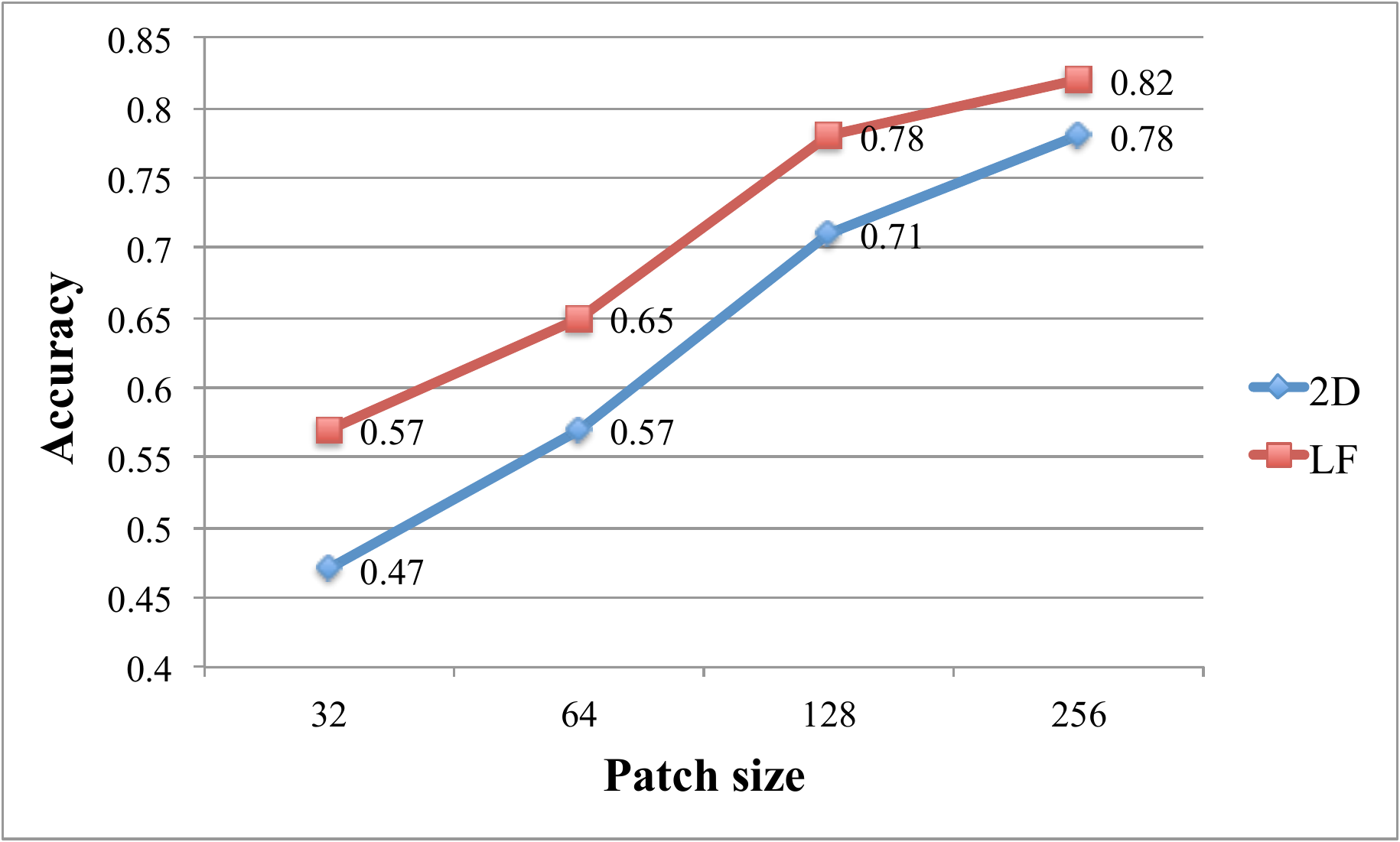} \label{fig: lf_gain:b}} \hfill\null
\caption{\textit{Prediction accuracy comparison for using 2D and light-field inputs. (a) We first show the accuracies for each category. It can be seen that using light-fields achieves the highest performance boost on leather, paper and wood, obtaining absolute gains of over 10\%. On the other hand, only the performance of glass drops. (b) Next, we vary the input patch size and test the performance again. It can be seen that as the patch size becomes smaller, the gain steadily increases.}}
\label{fig: lf_gain}
\end{figure}

\begin{figure}[t]
\captionsetup[subfigure]{labelformat=empty}
\centering 
\includegraphics[width=.85\linewidth]{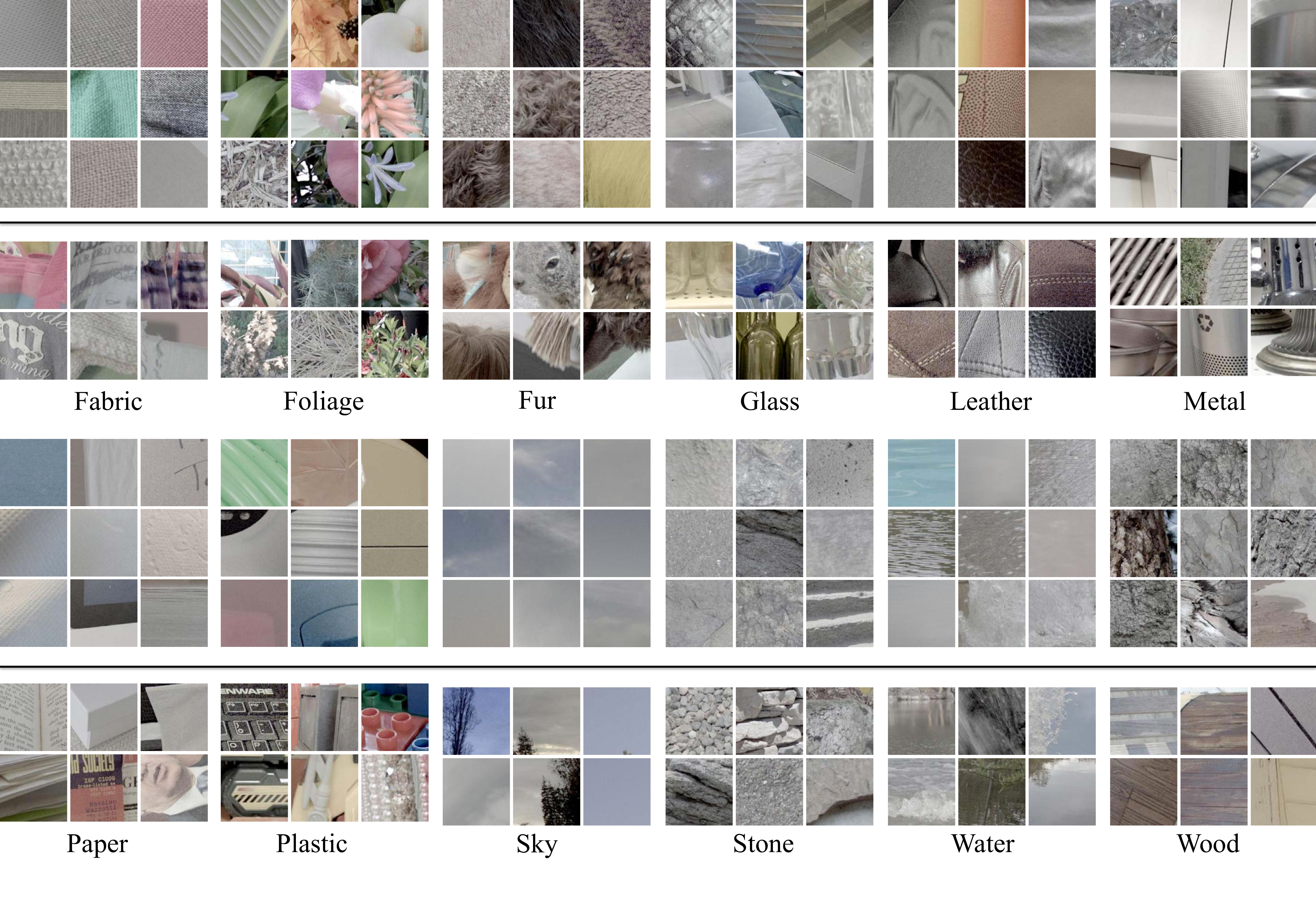}
\caption{\textit{Prediction result discrepancy between 2D and light-field inputs. The first $3\times 3$ grids show example patches that are predicted correctly using LF inputs, but misclassified using 2D inputs. The second grids show the opposite situation, where 2D models output the correct class but LF models fail. We found that the LF model performs the best when the object information is missing or vague, so we can only rely on the local texture, viewpoint change or reflectance information.}}
\label{fig:discrepancy}
\end{figure}

The confusion matrices for both 2D and light-field methods (using the angular filter method) are shown in Fig.~\ref{fig:confusion}, and a graphical comparison is shown in Fig.~\ref{fig: lf_gain:a}.
Relative to 2D images, using light-fields achieves the highest performance boost on leather, paper and wood, with absolute gains of over 10\%.
This is probably because the appearances of these materials are determined by complex effects such as subsurface scattering or inter-reflections, and multiple views help in disambiguating these effects.
Among all the 12 materials, only the performance for glass drops.
This is probably because the appearance of glass is often dependent on the scene rather than on the material itself.
Figure~\ref{fig:discrepancy} shows some examples that are misclassified using 2D inputs but predicted correctly using light-fields, and vice versa.
We observe that light-fields perform the best when the object information is missing or vague, necessitating reliance only on local texture or reflectance.
On the other hand, the 2D method often generates reasonable results if the object category in the patch is clear.

Next, we change the patch size for both methods, and test their accuracies to see the effect of patch size on performance gain.
We tried patch sizes 32, 64, 128 and 256 (Fig.~\ref{fig: lf_gain:b}).
It is observed that as we shrink the patch size from 128, the absolute gain steadily increases, from 7\% to 10\%.
If we look at the relative gain, it is growing even more rapidly, from about 10\% at size 128 to 20\% at size 32.
At size 256 the absolute gain becomes smaller.
A possibility is that at this scale, the object in the patch usually becomes apparent, and this information begins to dominate over the reflectance information.
Therefore, the benefits of light-fields are most pronounced when using small patches.
As the patch becomes smaller and smaller, it becomes harder and harder to recognize the object, so only local texture and reflectance information is available.
Also note that although increasing the patch size will lead to better accuracy, it will also reduce the output resolution for full scene classification, so it is a tradeoff and not always better.
Finally, while we have shown a significant increase in accuracy from 2D to 4D material recognition, once the dataset is published, our approach can still be improved by future advances that better exploit the full structure of 4D light-field data.

\subsection{Comparison between spatial/angular resolution} \label{sec:exp:spa}
Since light-field images contain more views, which results in an effectively larger number of pixels than 2D images, we also perform an experiment where the two inputs have the same number of pixels.
Specifically, we extract the light-field image with a spatial resolution of $128\times 128$ and an angular resolution of $4\times 4$, and downsample the image in the spatial resolution by a factor of 4.
This results in a light-field image of the same size as an original 2D image.
The classification results for using original 2D input and this downsampled light-field are 70.7\% and 75.2\% respectively.
Comparing with the original light-field results (77.8\%), we observe that reducing the spatial resolution lowers the prediction accuracy, but it still outperforms 2D inputs by a significant amount.

\subsection{Results on other datasets}
Finally, to demonstrate the generality of our model, we test it on other datasets.
Since no light-field material datasets are available, we test on the synthesized BTF database~\cite{weinmann2014material}.
The database captures a large number of different viewing and lighting directions on 84 instances evenly classified into 7 materials.
From the database we can render arbitrary views by interpolation on the real captured data.
We thus render light-field images and evaluate our model on these rendered images.

First, directly applying our model on the BTF database already achieves 65.2\% classification accuracy (for the materials that overlap).
Next, since the BTF database contains different material categories from our dataset, we use our models to extract the 4096-dimensional output of the penultimate fully connected layer.
This is the vector that is used to generate the final class probability in the network, and acts as a feature descriptor of the original input.
We then use this feature descriptor to train an SVM.
We pick two-thirds of the BTF dataset as training set and the rest as test set.
The results for using 2D and light-field inputs are 59.8\% and  63.7\% respectively.
Note that light-field inputs achieve about 4\% better performance than using 2D inputs.
Considering that the rendered images may not look similar to the real images taken with a  Lytro camera, this is a somewhat surprising result.
Next, we fine-tune our models on the training set, and test the performance on the test set again.
The results for using 2D and light-field inputs are 67.7\% and 73.0\% respectively.
Again using light-fields achieves more than 5\% performance boost.
These results demonstrate the generality of our models.

\subsection{Full scene material segmentation}
\begin{figure}[t]
\centering 
\subfloat{\includegraphics[width=.95\linewidth]{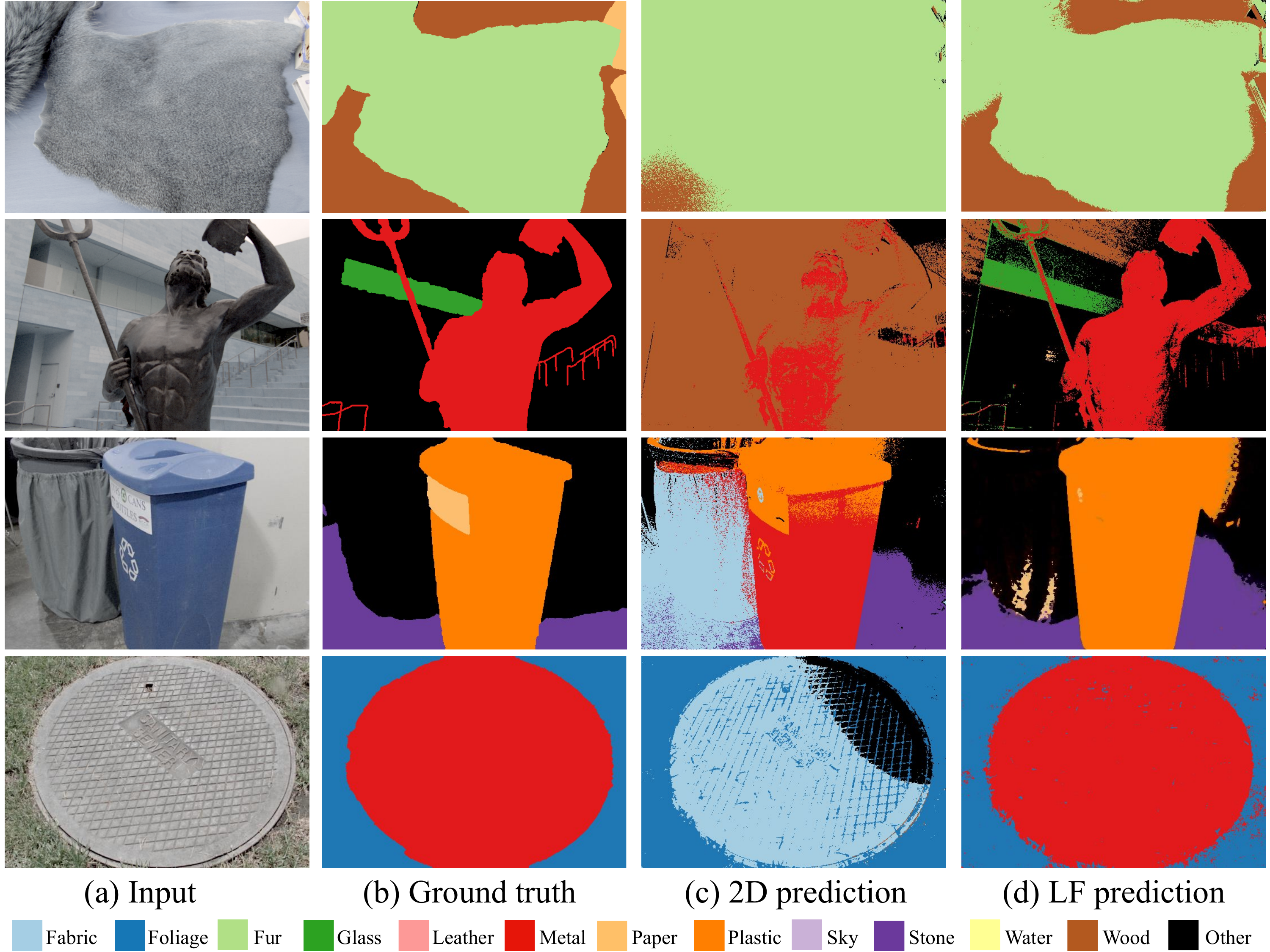}}
\caption{\textit{Full scene material classification examples. Bottom: legend for material colors. Compared with using 2D inputs, we can see that our light-field method produces more accurate prediction results. }}
\label{fig:scene}
\end{figure}

Finally, we convert our patch model to a fully convolutional model and test it on an entire image to perform material segmentation.
We do not directly train a fully convolutional network (FCN) since we find it very unstable and the training loss seldom converges.
Instead, we first train our model on image patches as described previously, convert it to a fully convolutional model, and then fine-tune it on entire images.
To train on a full image, we add another material class to include all other materials that do not fall into any of the 12 classes in our dataset.
We repeat this process for both our models of patch size 256 and 128 to get two corresponding FCN models, and combine their results by averaging their output probability maps.
Finally, as the probability map is low-resolution due to the network stride, we use edge-aware upsampling~\cite{he2010guided} to upsample the probability map to the same size as the original image.
The per pixel accuracy for FCN prediction before and after the guided filter is 77.0\% and 79.9\%, respectively.
The corresponding accuracies for 2D models are 70.1\% and 73.7\%, after we apply the same procedure.
Note that our method still retains 6-7\% boost compared with 2D models.
Example results for both methods are shown in Fig.~\ref{fig:scene}.

\section{Conclusion}
We introduce a new light-field dataset in this work.
Our dataset is the first one acquired with the Lytro Illum camera, and contains 1200 images, which is much larger than all previous datasets.
Since light-fields can capture different views, they implicitly contain the reflectance information, which should be helpful when classifying materials.
In view of this, we exploit the recent success in deep learning approaches, and train a CNN on this dataset to perform material recognition.
To utilize the pre-trained 2D models, we implement a number of different architectures to adapt them to light-fields, and propose a ``decomposed'' 4D filter.
These architectures provide insights to light-field researchers interested in adopting CNNs, and may also be generalized to other tasks involving light-fields in the future.
Our experimental results demonstrate that we can benefit from using 4D light-field images, obtaining an absolute gain of about 7\% in classification accuracy compared with using a single view alone.
Finally, although we utilize this dataset for material recognition, it can also spur research towards other applications that combine learning techniques and light-field imagery.

\sectionfont{\small}
\small
\section*{Acknowledgements}
This work was funded in part by ONR grant N00014152013, NSF grant IIS-1617234, Draper Lab, a Google Research Award, support by Nokia, Samsung and Sony to the UC San Diego Center for Visual Computing, and a GPU donation from NVIDIA.

\clearpage
\sectionfont{\normalsize}
\normalsize
\bibliographystyle{splncs03}
\bibliography{egbib}

\begin{thebibliography}{10}
\providecommand{\url}[1]{\texttt{#1}}
\providecommand{\urlprefix}{URL }

\bibitem{stanford}
Adams, A., Levoy, M., Vaish, V., Wilburn, B., Joshi, N.: Stanford light field
  archive, \url{http://lightfield.stanford.edu/}

\bibitem{bell2013opensurfaces}
Bell, S., Upchurch, P., Snavely, N., Bala, K.: Opensurfaces: A richly annotated
  catalog of surface appearance. ACM Transactions on Graphics (TOG)  32(4),
  111 (2013)

\bibitem{bell2015material}
Bell, S., Upchurch, P., Snavely, N., Bala, K.: Material recognition in the wild
  with the materials in context database. In: Proceedings of the IEEE
  Conference on Computer Vision and Pattern Recognition (CVPR) (2015)

\bibitem{caputo2005class}
Caputo, B., Hayman, E., Mallikarjuna, P.: Class-specific material
  categorisation. In: Proceedings of the IEEE International Conference on
  Computer Vision (ICCV) (2005)

\bibitem{cimpoi2014describing}
Cimpoi, M., Maji, S., Kokkinos, I., Mohamed, S., Vedaldi, A.: Describing
  textures in the wild. In: Proceedings of the IEEE Conference on Computer
  Vision and Pattern Recognition (CVPR) (2014)

\bibitem{cimpoi2015deep}
Cimpoi, M., Maji, S., Vedaldi, A.: Deep filter banks for texture recognition
  and segmentation. In: Proceedings of the IEEE Conference on Computer Vision
  and Pattern Recognition (CVPR) (2015)

\bibitem{cula20043d}
Cula, O.G., Dana, K.J.: {3D} texture recognition using bidirectional feature
  histograms. International Journal of Computer Vision  59(1),  33--60 (2004)

\bibitem{dana1999reflectance}
Dana, K.J., Van~Ginneken, B., Nayar, S.K., Koenderink, J.J.: Reflectance and
  texture of real-world surfaces. ACM Transactions on Graphics (TOG)  18(1),
  1--34 (1999)

\bibitem{farabet2013learning}
Farabet, C., Couprie, C., Najman, L., LeCun, Y.: Learning hierarchical features
  for scene labeling. IEEE Transactions on Pattern Analysis and Machine
  Intelligence (PAMI)  35(8),  1915--1929 (2013)

\bibitem{hayman2004significance}
Hayman, E., Caputo, B., Fritz, M., Eklundh, J.O.: On the significance of
  real-world conditions for material classification. In: Proceedings of the
  IEEE European Conference on Computer Vision (ECCV) (2004)

\bibitem{he2010guided}
He, K., Sun, J., Tang, X.: Guided image filtering. In: Proceedings of the IEEE
  European Conference on Computer Vision (ECCV) (2010)

\bibitem{hu2011toward}
Hu, D., Bo, L., Ren, X.: Toward robust material recognition for everyday
  objects. In: BMVC (2011)

\bibitem{jarabo2014people}
Jarabo, A., Masia, B., Bousseau, A., Pellacini, F., Gutierrez, D.: How do
  people edit light fields? ACM Transactions on Graphics (TOG)  33(4),  146--1
  (2014)

\bibitem{jia2014caffe}
Jia, Y., Shelhamer, E., Donahue, J., Karayev, S., Long, J., Girshick, R.,
  Guadarrama, S., Darrell, T.: Caffe: Convolutional architecture for fast
  feature embedding. In: Proceedings of the ACM International Conference on
  Multimedia (2014)

\bibitem{kim2013scene}
Kim, C., Zimmer, H., Pritch, Y., Sorkine-Hornung, A., Gross, M.H.: Scene
  reconstruction from high spatio-angular resolution light fields. ACM
  Transactions on Graphics (TOG)  32(4), ~73 (2013)

\bibitem{krizhevsky2012imagenet}
Krizhevsky, A., Sutskever, I., Hinton, G.E.: Imagenet classification with deep
  convolutional neural networks. In: Advances in neural information processing
  systems (2012)

\bibitem{li2014saliency}
Li, N., Ye, J., Ji, Y., Ling, H., Yu, J.: Saliency detection on light field.
  In: Proceedings of the IEEE Conference on Computer Vision and Pattern
  Recognition (CVPR) (2014)

\bibitem{li2012recognizing}
Li, W., Fritz, M.: Recognizing materials from virtual examples. In: Proceedings
  of the IEEE European Conference on Computer Vision (ECCV) (2012)

\bibitem{liu2010exploring}
Liu, C., Sharan, L., Adelson, E.H., Rosenholtz, R.: Exploring features in a
  bayesian framework for material recognition. In: Proceedings of the IEEE
  Conference on Computer Vision and Pattern Recognition (CVPR) (2010)

\bibitem{liu2014discriminative}
Liu, C., Gu, J.: Discriminative illumination: Per-pixel classification of raw
  materials based on optimal projections of spectral {BRDF}. IEEE Transactions
  on Pattern Analysis and Machine Intelligence (PAMI)  36(1),  86--98 (2014)

\bibitem{lombardi2012single}
Lombardi, S., Nishino, K.: Single image multimaterial estimation. In:
  Proceedings of the IEEE Conference on Computer Vision and Pattern Recognition
  (CVPR) (2012)

\bibitem{long2015fully}
Long, J., Shelhamer, E., Darrell, T.: Fully convolutional networks for semantic
  segmentation. In: Proceedings of the IEEE Conference on Computer Vision and
  Pattern Recognition (CVPR) (2015)

\bibitem{marwah2013compressive}
Marwah, K., Wetzstein, G., Bando, Y., Raskar, R.: Compressive light field
  photography using overcomplete dictionaries and optimized projections. ACM
  Transactions on Graphics (TOG)  32(4),  1--11 (2013)

\bibitem{nicodemus1977geometrical}
Nicodemus, F.E., Richmond, J.C., Hsia, J.J., Ginsberg, I.W., Limperis, T.:
  Geometrical considerations and nomenclature for reflectance, vol. 160. US
  Department of Commerce, National Bureau of Standards Washington, DC, USA
  (1977)

\bibitem{oquab2014learning}
Oquab, M., Bottou, L., Laptev, I., Sivic, J.: Learning and transferring
  mid-level image representations using convolutional neural networks. In:
  Proceedings of the IEEE Conference on Computer Vision and Pattern Recognition
  (CVPR) (2014)

\bibitem{qi2014pairwise}
Qi, X., Xiao, R., Li, C.G., Qiao, Y., Guo, J., Tang, X.: Pairwise rotation
  invariant co-occurrence local binary pattern. IEEE Transactions on Pattern
  Analysis and Machine Intelligence (PAMI)  36(11),  2199--2213 (2014)

\bibitem{raghavendra2016exploring}
Raghavendra, R., Raja, K.B., Busch, C.: Exploring the usefulness of light field
  cameras for biometrics: An empirical study on face and iris recognition. IEEE
  Transactions on Information Forensics and Security  11(5),  922--936 (2016)

\bibitem{schwartz2013visual}
Schwartz, G., Nishino, K.: Visual material traits: Recognizing per-pixel
  material context. In: Proceedings of the IEEE International Conference on
  Computer Vision (ICCV) Workshops (2013)

\bibitem{sharan2009material}
Sharan, L., Rosenholtz, R., Adelson, E.: Material perception: What can you see
  in a brief glance? Journal of Vision  9(8),  784--784 (2009)

\bibitem{simonyan2014very}
Simonyan, K., Zisserman, A.: Very deep convolutional networks for large-scale
  image recognition. arXiv preprint arXiv:1409.1556  (2014)

\bibitem{su2015multi}
Su, H., Maji, S., Kalogerakis, E., Learned-Miller, E.: Multi-view convolutional
  neural networks for 3d shape recognition. In: Proceedings of the IEEE
  International Conference on Computer Vision (ICCV) (2015)

\bibitem{szegedy2015going}
Szegedy, C., Liu, W., Jia, Y., Sermanet, P., Reed, S., Anguelov, D., Erhan, D.,
  Vanhoucke, V., Rabinovich, A.: Going deeper with convolutions. In:
  Proceedings of the IEEE Conference on Computer Vision and Pattern Recognition
  (CVPR) (2015)

\bibitem{tao2013depth}
Tao, M.W., Hadap, S., Malik, J., Ramamoorthi, R.: Depth from combining defocus
  and correspondence using light-field cameras. In: Proceedings of the IEEE
  International Conference on Computer Vision (ICCV) (2013)

\bibitem{wang2015occlusion}
Wang, T.C., Efros, A., Ramamoorthi, R.: Occlusion-aware depth estimation using
  light-field cameras. In: Proceedings of the IEEE International Conference on
  Computer Vision (ICCV) (2015)

\bibitem{wanner2013datasets}
Wanner, S., Meister, S., Goldl{\"u}cke, B.: Datasets and benchmarks for densely
  sampled {4D} light fields. In: Annual Workshop on Vision, Modeling and
  Visualization. pp. 225--226 (2013)

\bibitem{weinmann2014material}
Weinmann, M., Gall, J., Klein, R.: Material classification based on training
  data synthesized using a {BTF} database. In: Proceedings of the IEEE European
  Conference on Computer Vision (ECCV) (2014)

\bibitem{yoon2015learning}
Yoon, Y., Jeon, H.G., Yoo, D., Lee, J.Y., Kweon, I.: Learning a deep
  convolutional network for light-field image super-resolution. In: Proceedings
  of the IEEE International Conference on Computer Vision (ICCV) Workshops
  (2015)

\bibitem{zhang2015reflectance}
Zhang, H., Dana, K., Nishino, K.: Reflectance hashing for material recognition.
  In: Proceedings of the IEEE Conference on Computer Vision and Pattern
  Recognition (CVPR) (2015)

\end{thebibliography}
\end{document}